\pdfoutput=1

\documentclass[11pt]{article}

\usepackage{ACL2023}

\usepackage{times}
\usepackage{latexsym}
\usepackage{amsmath}
\usepackage{tcolorbox}
\usepackage{multirow}
\usepackage{xcolor}

\usepackage[T1]{fontenc}

\usepackage[utf8]{inputenc}

\usepackage{microtype}

\usepackage{inconsolata}

\title{Scaling Down Semantic Leakage: Investigating Associative Bias in Smaller Language Models}

\author{Veronika Smilga \\
  University of Tübingen \\
  \texttt{smilgaveronika@gmail.com} \\}

\begin{document}
\maketitle
\begin{abstract}
Semantic leakage is a phenomenon recently introduced by~\citet{gonen2024does}. It refers to a situation in which associations learnt from the training data emerge in language model generations in an unexpected and sometimes undesired way. Prior work has focused on leakage in large language models (7B+ parameters). In this study, I use Qwen2.5 model family to explore whether smaller models, ranging from 500M to 7B parameters, demonstrate less semantic leakage due to their limited capacity for capturing complex associations. Building on the previous dataset from~\citet{gonen2024does}, I introduce a new dataset of color-focused prompts, categorized into specific types of semantic associations, to systematically evaluate the models' performance. Results indicate that smaller models exhibit less semantic leakage overall, although this trend is not strictly linear, with medium-sized models sometimes surpassing larger ones in leaking behavior. The dataset, the model generations, and the evaluation code are publicly available at \url{https://github.com/smilni/semantic_leakage_project}.
\end{abstract}

\section{Introduction}
\label{intro}
Language Models (LMs) have become omni-present in our everyday lives. However, the way they function is still largely a black-box. Inherently influenced by the training data, LMs demonstrate human-like perception of right and wrong~\citep{DBLP:journals/corr/abs-2103-11790}, while also exhibiting a variety of undesirable social biases ~\citep{bolukbasi2016mancomputerprogrammerwoman, lucy-bamman-2021-gender}. ~\citet{10.1145/3597307} argue that the source of such biases are associations learnt by the model in the course of its training. 

A broader example of learnt associations surfacing in an unexpected and unsuitable way is \textbf{semantic leakage}~\citep{gonen2024does}. The association learnt by the model might make a concept from a prompt influence the model generation out of proportion. That results in bizarre outputs, such as \textit{``The dinner was served on pink plates. Today's dish was... \textbf{rose petal soup}''}\footnote{Generated by Qwen2.5-3B-Instruct}. Another interesting case of semantic leakage is the leakage of literal word-by-word meaning of idioms and proper names, such as \textit{``Ivory is a student. Her favorite color is... \textbf{white}''}\footnote{Generated by Qwen2.5-1.5B-Instruct}.

Semantic leakage in LMs is closely related to problems in word-to-concept mapping in image generation. With prompts containing several objects, the attributes of the objects may leak into each other, get interchanged, or one of the objects may be omitted from the image altogether~\citep{feng2023trainingfreestructureddiffusionguidance, dahary2024yourselfboundedattentionmultisubject}. Conversely, the presence of one ambiguous word in a prompt may lead to the model producing a picture of several entities corresponding to each of the word's meanings~\citep{rassin2022dalle2seeingdoubleflaws}.

While the above-described problems with word-to-concept mapping problems lead to errors in image generation, one may argue that semantic leakage in text generation, on the opposite, helps in maintaining the coherence of the text. Both semantic leakage and bias in machine learning models seem to be rooted in undesirable associations learnt during training. Unlike inherent model bias, semantic leakage does not necessarily result in incorrect or harmful generations. However, as pointed out by~\citet{gonen2024does}, semantic leakage is a common pattern that characterises language models across different sizes and architectures, large and small, instruction-tuned and general-purpose ones. It would thus be interesting to investigate the phenomenon further to understand how models internalize and use associations.

~\citet{gonen2024does} analyse semantic leakage in various categories with strong semantic association component, including colors, food, animals, songs, and occupations. They test several Large LMs, including models of Llama-2 7B+ and LLama-3 8B+ families as well as OpenAI's GPT3.5, GPT4 and GPT4o. 

In this project, I further investigate the phenomenon of semantic leakage. Building on~\citet{gonen2024does} who examined models of 7 billion parameters and larger, I choose to focus on smaller LMs, from 500 million to 7 billion parameters. The goal is to determine whether the model's size affects its proneness to semantic leakage. I hypothesize that smaller models, due to their size, may be unable to capture the semantic associations that result in semantic leakage. Thus, I am expecting smaller LMs to exhibit less leaking behavior as compared to larger ones.

Additionally, I build a new set of prompts focusing on the category of color and test the models both on the original dataset by~\citet{gonen2024does} and the color-focused one. Color-related prompts are divided into three categories: 1) prompts with a mention of a color, with a non-color-related concept to be generated; 2) prompts with a mention of a color, with another color to be generated; 3) prompts with names or set expressions mentioning a color, with another color to be generated\footnote{Refer to Subsection~\ref{sub:datasets} for example prompts of each category.}. The goal is to test whether the models exhibit leakage more for any of the categories. I hypothesize that prompts containing a color with another color to be generated, i.e. prompts of type 2) and type 3), leak more as models may take short-cuts by repeating previously mentioned colors.

\section{Methods}
\subsection{Models}
Qwen2.5~\citep{qwen2.5} is a series of large language models sized from 500 million to 72 billion parameters, both general-purpose and instruction-fine-tuned. Qwen2.5 is the most capable state-of-the-art series of models with open weights and smaller counterparts available as of today\footnote{As of October 1st, 2024, according to \href{https://lmarena.ai}{Chatbot Arena LLM Leaderboard}~\citep{chiang2024chatbot}, among models with open weights, Qwen2.5-72b-Instruct is only surpassed by Meta-Llama-3.1-405b-Instruct. However, Llama-3.1 model family only features models of 8B, 70B, and 405B parameters, while I aim to evaluate models of 7B and smaller.}. I test all Qwen2.5-Instruct models up to 7B parameters: Qwen2.5-0.5B-Instruct, Qwen2.5-1.5B-Instruct, Qwen2.5-3B-Instruct, Qwen2.5-7B-Instruct-GPTQ-Int4\footnote{Dut to computational constraints, I had to use GPTQ-quantised~\citep{frantar2023gptqaccurateposttrainingquantization} version of Qwen2.5-7B-Instruct that would fit into a single P100 GPU.}. Only Instruct models are tested as, according to~\citet{gonen2024does}, instruction-finetuned models exhibit more leaking behavior than their general-purpose counterparts. In addition to that, I found it hard to get meaningful generations from smaller non-instruction-finetuned Qwen models.

All models were accessed using Huggingface\footnote{\url{https://huggingface.com}}. Inference was run using Kaggle notebooks\footnote{\url{https://www.kaggle.com}} on a single P100 GPU. Temperature was set to an intermediate value of 0.5 to ensure balance between exploring diverse token probabilities and preserving high-probability sequences. Maximum number of tokens to be generated was set to 10. Additionally, generated text was capped after the first full stop, exclamation or question mark, following the original paper. 

\subsection{Evaluation metrics}
\label{evals}

\begin{table*}[t!]
    \centering
    \begin{tabular}{p{1cm}|p{3cm}|p{3cm}|p{7cm}}
        \textbf{Cat.} & \textbf{Prompt type} & \textbf{Generation type} & \textbf{Example prompt \& generation} \\
        \hline
        1 & Color & Non-color-related item & He likes \textbf{red}. He works as a... \textcolor{teal}{\textbf{firefighter}} \\
        \hline
        2 & Color & Color & \textbf{White} clouds floated in the summer sky. The fence was painted... \textcolor{teal}{\textbf{white and weathered}} \\
        \hline
        3 & Color-related name or set expression & Color & Mary saw the world \textbf{through rose-colored glasses}. She was wearing a T-shirt colored... \textcolor{teal}{\textbf{pink}} \\
    \end{tabular}
    \caption{Dataset categories. Color-related concepts in prompt are in black bold. Color-related concepts in model generation are in green bold. All example generations were obtained from Qwen2.5-7B-Instruct.}
    \label{tab:prompt-categories}
\end{table*}

\begin{table*}[h]
\centering
\begin{tabular}{|c|c|c|c|}
\hline
\textbf{Dataset} & \textbf{Model} & \textbf{BERT-score} & \textbf{SentenceBERT}\\
\hline
\multirow{4}{*}{\shortstack{Various semantic categories \\ \cite{gonen2024does}}} & 0.5B-Instruct       & 69.27              & 77.52\\
                        & 1.5B-Instruct        & 75.23             & 79.36\\
                        & 3B-Instruct      & \textbf{83.03}              & 73.85\\
                        & 7B-Instruct-GPTQ-Int4      & 74.77            & \textbf{80.64}\\
\hline
\multirow{4}{*}{\shortstack{Color-related \\ (the new dataset)}} & 0.5B-Instruct       & 60.68              & 61.23\\
                        & 1.5B-Instruct        & 71.77              & 69.58\\
                        & 3B-Instruct      & \textbf{77.41}              & \textbf{80.33}\\
                        & 7B-Instruct-GPTQ-Int4      & 73.41          & 66.39\\
\hline
\end{tabular}
\caption{Mean Leak-Rate for original and new color-related dataset, Qwen2.5 model family. The highest score for each dataset and score type across the model sizes is in bold.}
\label{tab:leak-rate}
\end{table*}

To evaluate how much a model exhibit semantic leakage, I use Mean Leak-Rate suggested by~\citet{gonen2024does}. This metric operates by comparing similarity of a concept that is supposed to trigger semantic leakage, a control generation for a prompt without the concept and a test generation for a prompt containing the concept. Let us examine the example from \ref{intro}, \textit{``Ivory is a student. Her favorite color is... \textbf{white}''}. The leakage-triggering concept is \textit{Ivory}, a name that is likely to be associated with color white due to the original meaning of the word \textit{ivory}. The test prompt leading to test generation is \textit{``Ivory is a student. Her favorite color is''}, and the control prompt leading to control generation is \textit{``Her favorite color is''}. Test generation \textit{white} is an example of leaking behavior. 

Leak-Rate is the percentage of instances when the similarity of the concept and the test generation is higher than that of the concept and the control generation. Leak-Rate for one prompt is calculated as follows:

\begin{align}
\label{eq:leak-rate}
\resizebox{\columnwidth}{!}{
$
\text{Leak-Rate} =
\begin{cases}
100 & \text{if } sim(\text{concept}, \text{test}) > sim(\text{concept}, \text{control}) \\
0 & \text{if } sim(\text{concept}, \text{test}) < sim(\text{concept}, \text{control}) \\
50 & \text{if } sim(\text{concept}, \text{test}) = sim(\text{concept}, \text{control}) 
\end{cases}
$
}
\tag{~\citealt{gonen2024does}}
\end{align}

To obtain the final Mean Leak-Rate of a model, the Leak-Rate is then averaged across all test prompts:
\begin{align}
\text{Mean Leak-Rate} &= \frac{1}{N} \sum_{i=1}^{N} \text{Leak-Rate\_i}
\end{align}

where \( \text{Leak-Rate\_i} \) is the Leak-Rate for prompt \( i \) and \( N \) is the total number of prompts.

To compute similarity between concepts and generations, I use \texttt{BERT-score} \cite{bertscore} with \texttt{distilbert-base-uncased}\footnote{\url{https://huggingface.co/distilbert/distilbert-base-uncased}} as the backbone model and \texttt{SentenceBERT} 
\cite{sentence-bert} with \texttt{all-MiniLM-L6-v2}\footnote{\url{https://huggingface.co/sentence-transformers/all-MiniLM-L6-v2}} as the backbone model. Since the original paper's ~\citep{gonen2024does} code is not yet available, I implement the calculations myself, following the description provided by the authors.

\subsection{Datasets}
\label{sub:datasets}
\begin{table*}[h]
\centering
\begin{tabular}{|c|c|c|c|}
\hline
\textbf{Cat.} & \textbf{Model} & \textbf{BERT-score} & \textbf{SentenceBERT}\\
\hline

\multirow{4}{*}{\shortstack{1) Color in prompt, \\ other concept expected in generation}} & 0.5B-Instruct       & 65.03              & \textbf{70.61}\\
                        & 1.5B-Instruct        & \textbf{73.18}              & \textbf{72.79}\\
                        & 3B-Instruct      & 70.45              & 73.76\\
                        & 7B-Instruct-GPTQ-Int4      & \textbf{88.45}            & \textbf{70.3}\\
\hline
\multirow{4}{*}{\shortstack{2) Color in prompt, \\ color expected in generation}} & 0.5B-Instruct       & 56.33              & 51.85\\
                        & 1.5B-Instruct        & 70.36              & 66.36\\
                        & 3B-Instruct      & \textbf{84.36}             & \textbf{86.91}\\
                        & 7B-Instruct-GPTQ-Int4      & 58.36          & 62.48\\
\hline
\multirow{4}{*}{\shortstack{3) Name or set expression in prompt, \\ color expected in generation}} & 0.5B-Instruct       & \textbf{71.22}              & \textbf{70.61}\\
                        & 1.5B-Instruct        & 66.33              & 61.02\\
                        & 3B-Instruct      & 57.76              & 72.45\\
                        & 7B-Instruct-GPTQ-Int4      & 60.82          & 64.9\\
\hline
\end{tabular}
\caption{Mean Leak-Rate for different categories of color-related prompts, Qwen2.5 model family. The highest score for each model size across all categories is in bold.}
\label{tab:cats}
\end{table*}
As a starting point, I evaluate the models' proneness to semantic leakage on the initial dataset of 109 prompts related to various semantic categories from~\citet{gonen2024does}. In order to calculate Leak-Rate on the original dataset, I manually identify the concept and the corresponding control item for each test prompt.

Additionally, I build a new extended dataset of prompts related to colors. This dataset consists of three parts: 1) prompts with color as the concept and an expected non-color-related generation; 2) prompts with color as the concept and an expected color-related generation; 3) prompts with color-related name or set expression as the concept and an expected color-related generation. See Table~\ref{tab:prompt-categories} for an example of a prompt and a generation for each category.

To construct prompts for categories 1) and 2), I handcraft 20 templates (10 for each category), such as \textit{``The flowerpot is \{INSERT\}. The flower is a ...''}. I make sure that in each case any color can be inserted to replace the placeholder \textit{\{INSERT\}}, i.e. none of the templates are tied to specific colors. To measure semantic leakage as described in Subsection~\ref{evals}, each template has a corresponding control prompt, e.g. \textit{``The flower is a ...''}. 

Each template is then completed using 11 basic color terms, resulting in 220 prompts. Following~\citet{berlin1991basic}, I define basic color terms for English as black, white, red, green, yellow, blue, brown, orange, pink, purple, and gray. To further diversify the dataset, I also include two more specific shades or variations for each basic color, resulting in the final dataset of 660 prompts for categories 1) and 2). You may find the full list of colors used to complete the templates in Appendix~\ref{appendix:a}. 

To construct prompts for category 3), I handcraft 4 templates, such as \textit{``\{INSERT\} is a student. Her favorite color is ...''} for names and \textit{``\{INSERT\} He was wearing a T-shirt colored ...''} for set expressions. Same as for other categories, each template has a corresponding control prompt. 

When constructing prompts for category 3), simple slot-filling was not always sufficient. Prompts featuring set expressions were created manually to ensure that the result was coherent and the set expression was contextually relevant. The final dataset contains 60 prompts of category 3), each featuring a unique color-rated proper name or expression. You may find the full list of color-related names and set expression used to complete the templates in Appendix~\ref{appendix:b}. 

\section{Results}

Table~\ref{tab:leak-rate} presents the resulting Leak-Rate scores calculated with the use of BERT-score and SentenceBERT for the four models of Qwen2.5-Instruct family for the original and the new dataset. For all models and datasets, the Leak-Rate exceeds 50\%, meaning that test generations show higher semantic similarity to the corresponding concepts in comparison with control generations. A 50\% Leak-Rate would indicate random similarity and no leakage, while values above 50\% confirm that semantic leakage is present in the models' output.

Models of all sizes demonstrate consistently high Leak-Rate on the original dataset from~\citet{gonen2024does}. Interestingly, compared to Llama and OpenAI models evaluated in~\citet{gonen2024does}, Qwen2.5 model family appears to be even more prone to leakage. Qwen2.5-3B-Instruct achieves the highest BERT-score-based Mean Leak-Rate of 83.03, surpassing the highest BERT-score-based Mean Leak-Rate of 78.1 reported for Llama-3-8B-Instruct by~\citet{gonen2024does}. Similarly, Qwen2.5-7B-Instruct-GPTQ-Int4 demonstrates the highest SentenceBERT-based Mean Leak-Rate of 80.64, compared to the highest score of 71.7 reported for Llama-2-7B-Chat in the original paper.

Larger models seem to demonstrate higher Leak-Rate, with Qwen2.5-3B-Instruct and Qwen2.5-7B-Instruct-GPTQ-Int4 having the highest BERT-score-based and SentenceBERT-based Mean Leak-Rate score for both original dataset and newly constructed color-related dataset. Interestingly, the smallest model, Qwen2.5-0.5B-Instruct, seems to demonstrate the least leaking behavior on both datasets in terms of both BERT-score-based and SentenceBERT-based Mean Leak-Rate. 

The hypothesis that smaller language models leak less information than larger ones appears to hold, at least partially. Indeed, Qwen2.5-0.5B-Instruct is the smallest model of the family, and its Mean-Leak Rate is consistently the lowest or next to the lowest. However, Qwen2.5-7B-Instruct-GPTQ-Int4 is the largest model of the family included in my analysis, but in three out of four cases it demonstrates lower Mean Leak-Rate than Qwen2.5-3B-Instruct. I would generalize that within the Qwen2.5 family models under 7 billion parameters tend to exhibit more leakage as their size increases, but this relationship is not strictly linear – in most cases, the highest Mean Leak-Rate score is demonstrated by the second-largest model.

Table~\ref{tab:cats} presents the BERT-score-based and SentenceBERT-based Leak-Rate scores for each of the categories of the color-related dataset. Models of different sizes demonstrate distinct behaviors in terms of the degree of semantic leakage across different categories. However, after averaging the Mean Leak-Rate for each category, it is evident that in general models tend to leak more for prompts belonging to category 1), with mean BERT-score-based LR of 74.28 and SentenceBERT-based LR of 71.87, as compared to category 2), with mean BERT-score-based LR of 67.35 and SentenceBERT-based LR of 66.90, and category 3), with mean BERT-score-based LR of 64.03 and SentenceBERT-based LR of 67.25. 

The hypothesis that prompts of categories 2) and 3), i.e. containing a color and leading to generation of another color, would lead to more leaking behavior, does not generally hold true. It is only true for Qwen2.5-3B-Instruct because this model tends to repeat parts of the prompt in the generation, as I will further demonstrate in Section~\ref{disc}. Contexts like those in 2) and 3) make concept repetition particularly convenient as the same kind of concept is featured in the prompt and expected to be generated. However, in general, models, including Qwen2.5-1.5B-Instruct and Qwen2.5-7B-Instruct-GPTQ-Int4 tend to leak more for prompts of category 1), i.e. prompts with a concept other than a color to be generated.

\section{Discussion}
\label{disc}

It is important to note that high Mean Leak-Rate of Qwen2.5-3B-Instruct is mostly explained by the fact that this model tended to repeat the concept mentioned in the prompt in its generations, e.g. \textit{``The cake was topped with tangerine orange icing. The candles were colored... \textbf{tangerine orange}''}. None of the other models, smaller or larger, exhibit repetitions to such degree. All models were run with the same hyperparameters, so I find this tendency hard to explain. However, it is important to take this peculiarity of Qwen2.5-3B-Instruct into account when drawing conclusion about the leaking behaviour of the models.

The lower Mean Leak-Rate of Qwen2.5-0.5B-Instruct and Qwen2.5-1.5B-Instruct may be explained by the fact that these models tend to generate the same continuations to the prompts regardless of the preceding context. For example, in \textit{``The house was painted charcoal black / emerald green / chestnut brown / coral orange. She saw a... \textbf{tree in the yard}''}\footnote{Generated by Qwen2.5-0.5B-Instruct}, the generation is the same for different colors in prompt. In other words, the immediate context in the end of the prompt seems to affect the model more than the potential leakage triggering concept. 

That does not happen with larger models – both Qwen2.5-3B-Instruct and Qwen2.5-7B-Instruct-GPTQ-Int4 seem to be much more sensitive to the context, with few to none repeated generations for different prompt variations. For example, for the above-mentioned template, there each color triggers a different generation: \textit{``The house was painted charcoal black / emerald green / chestnut brown / coral orange. She saw a... \textbf{ghostly figure / shimmering hue / squirrel on the roof / sudden surprise}''}\footnote{Generated by Qwen2.5-7B-Instruct}

It seems that semantic leakage does indeed stem from the associations learnt by the model during training. In cases like the one described above, semantic leakage adds to the lexical richness and diversity of the generated text. Small models, such as Qwen2.5-0.5B-Instruct, exhibit less leaking behavior, but at the same time provide less diverse and less context-aware generations. In other words, it is important to understand that semantic leakage is not inherently negative but rather a byproduct of the associative knowledge learnt by the model. While undesirable in certain scenarios, such as those requiring unbiased or contextually accurate outputs, semantic leakage can make a positive contribution in other cases.

One of the limitations of the current study is the way semantic leakage is measured. In some cases, the repetition of the concept from the prompt in the generation is justified as it is used to refer to a previously mentioned concept, e.g. \textit{``The house was painted green. She saw a... \textbf{green house}''}\footnote{Generated by Qwen2.5-7B-Instruct-GPTQ-Int4}. However, generations like that are definitely semantically related to the concept of \textit{green} from the prompt and thus get classified as instances of semantic leakage, unjustly increasing the Mean Leak-Rate of the model.

Another limitation of the current study is the imbalance of categories in the new color-related dataset, with 330 prompts of category 1), 330 prompts of category 2), and only 60 prompts of category 3). This situation is explained by the fact that prompts of categories 1) and 2) were produced semi-automatically, by slot-filling colors in pre-written templates, while prompts of category 3) were crafted manually in an attempt to preserve coherence in prompts featuring set expressions. However, such imbalance makes our conclusions about the differences in semantic leakage for each of the categories less reliable.

Possible directions for future work include addressing the category imbalance. The smallest category 3) can be expanded by using the same names and set expressions in more than one prompt. Additionally, one may want to test larger models of the same family, Qwen2.5-14B-Instruct, Qwen2.5-32B-Instruct, and Qwen2.5-72B-Instruct, to determine whether the tendency of larger models to leak more holds for 14B+ models. Finally, it may be interesting to investigate the ways of detecting and mitigating undesirable leaking behavior in language models.

\bibliography{custom}
\bibliographystyle{acl_natbib}

\appendix
\onecolumn

\section{Basic colors and their variations, dataset categories a) and b)}
\label{appendix:a}

\begin{table}[h!]
\centering
\begin{tabular}{|p{2cm}|p{10cm}|}
\hline
\textbf{Basic color} & \textbf{Variations} \\ \hline
yellow & golden yellow, lemon yellow \\ \hline
red    & crimson red, scarlet red \\ \hline
purple & violet purple, lavender purple \\ \hline
green  & emerald green, olive green \\ \hline
blue   & navy blue, cobalt blue \\ \hline
pink   & fuchsia pink, blush pink \\ \hline
black  & charcoal black, jet black \\ \hline
white  & ivory white, snow white \\ \hline
brown  & chestnut brown, mahogany brown \\ \hline
orange & coral orange, tangerine orange \\ \hline
gray   & slate gray, ash gray \\ \hline
\end{tabular}
\label{tab:color-variations}
\end{table}

\section{Color-related names and set expressions, dataset category c)}
\label{appendix:b}

\begin{table}[h!]
\centering
\begin{tabular}{|p{2cm}|p{10cm}|}
\hline
\textbf{Type} & \textbf{Example} \\ \hline
Names & Rose, Amber, Ruby, Scarlett, Violet, Jade, Hazel, Indigo, Ivory, Coral, Sienna, Olive, Ebony, Gray, Pearl \\ \hline
Toponyms & Yellowknife (Canada), Redmond (Washington), Purple Springs (Alberta), Greenwich (UK), Bluefield (West Virginia), Roseland (Chicago), Blackfoot (Idaho), White Plains (New York), Brownsville (Texas), Orange (France), Grayling (Michigan) \\ \hline
Set expressions & to be a black sheep, to be tickled pink, to be in the red, to tell a white lie, to come out of the blue, to see a golden opportunity, to be in a gray area, to get the green light, to paint the town red, to have a silver tongue, to have a rosy outlook, to be as gray as the cloudy sky, to be in a black mood, to think in black and white, to have white-hot anger, to feel like a green rookie, to have rosy cheeks, to be in a brown study, to see the world through rose-colored glasses \\ \hline
\end{tabular}
\label{tab:color-variations}
\end{table}

\end{document}